\documentclass[10pt,twocolumn,letterpaper]{article}

\usepackage[pagenumbers]{cvpr} 

\usepackage{graphicx}
\usepackage{amsmath}
\usepackage{amssymb}
\usepackage{booktabs}
\usepackage{siunitx}

\usepackage[pagebackref,breaklinks,colorlinks]{hyperref}
\usepackage{multirow}
\usepackage{graphicx}

\usepackage{background}

\newcommand{\mywatermark}{%
    \begin{minipage}{\textwidth}
        \centering
        \fontsize{10}{10}\selectfont 
        This paper has been accepted at the CVPR 2024 Workshop on Synthetic Data for Computer Vision (SynData4CV) 
    \end{minipage}%
}

\backgroundsetup{
    scale=1,
    color=gray,
    angle=0,
    opacity=0.5,
    position=current page.north,
    vshift=-1cm, 
    contents={\mywatermark}
}

\usepackage[capitalize]{cleveref}
\crefname{section}{Sec.}{Secs.}
\Crefname{section}{Section}{Sections}
\Crefname{table}{Table}{Tables}
\crefname{table}{Tab.}{Tabs.}

\def\cvprPaperID{47} 
\def\confName{CVPR}
\def\confYear{2024}

\begin{document}

\title{SEVD: Synthetic Event-based Vision Dataset for Ego and Fixed Traffic Perception}

\author{Manideep Reddy Aliminati $^*$, Bharatesh Chakravarthi$^*$, Aayush Atul Verma, \\ Arpitsinh Vaghela, Hua Wei, Xuesong Zhou, Yezhou Yang \\
Arizona State University \\
{\tt\small \{areddy27, bshettah, averma90, avaghel3, hua.wei, xzhou74, yz.yang\}@asu.edu}\
}


\twocolumn[{%
\renewcommand\twocolumn[1][]{#1}%
\maketitle
\begin{center}
\captionsetup{type=figure}
\includegraphics[width=0.98\textwidth]{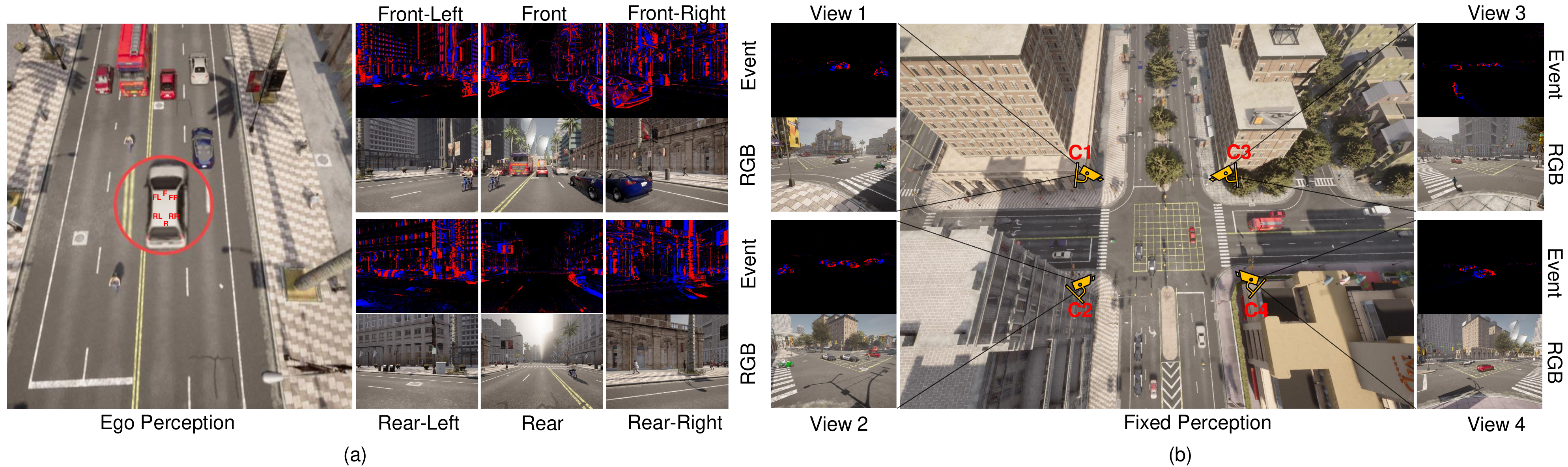}
\captionof{figure}{Bird's Eye View of Ego and Fixed Perception Scenario: (a) Shows the six views (Front-Left, Front, Front-Right, Rear-Left, Rear, Rear-Right) from an ego vehicle perception (highlighted in red circle) depicted through event-based and its corresponding RGB frames.  (b) Shows four views of an intersection from fixed cameras (C1, C2, C3, C4), with event-based and RGB frames for each view.}
\label{fig01}
\end{center}%
}]

\begin{figure*}[ht]
\centering
\includegraphics[width=0.92\linewidth]{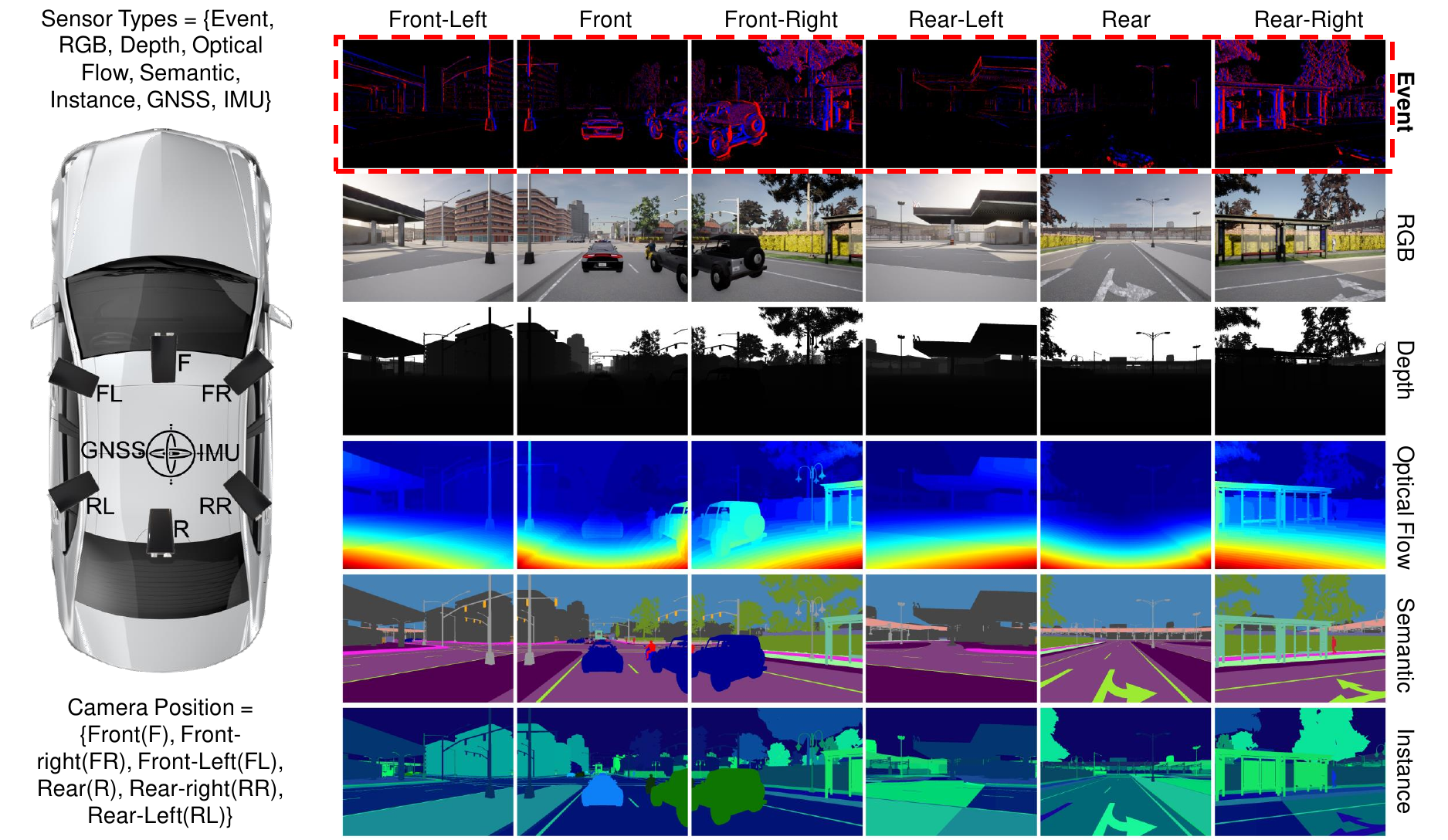}
\caption{Navigating the Dynamic Landscape of Road Traffic: A glimpse of ego perception data offering six views through event-based vision supported by RGB, depth, optical-flow, semantic, and instance segmentation sensor data generated using CARLA.}
\label{fig02}
\end{figure*}

\def\thefootnote{*}\footnotetext{Equal contribution}

\begin{abstract}
Recently,  event-based vision sensors have gained attention for autonomous driving applications, as conventional RGB cameras face limitations in handling challenging dynamic conditions. However, the availability of real-world and synthetic event-based vision datasets remains limited. In response to this gap, we present \textit{SEVD}, a first-of-its-kind multi-view ego, and fixed perception synthetic event-based dataset using multiple dynamic vision sensors within the CARLA simulator. Data sequences are recorded across diverse lighting (noon, nighttime, twilight) and weather conditions (clear, cloudy, wet, rainy, foggy) with domain shifts (discrete and continuous). \textit{SEVD} spans urban, suburban, rural, and highway scenes featuring various classes of objects (car, truck, van, bicycle, motorcycle, and pedestrian). Alongside event data, \textit{SEVD} includes RGB imagery, depth maps, optical flow, semantic, and instance segmentation, facilitating a comprehensive understanding of the scene. Furthermore, we evaluate the dataset using state-of-the-art event-based (RED, RVT) and frame-based (YOLOv8) methods for traffic participant detection tasks and provide baseline benchmarks for assessment. Additionally, we conduct experiments to assess the synthetic event-based dataset's generalization capabilities. The dataset is available at \url{https://eventbasedvision.github.io/SEVD}
\end{abstract}

\section{Introduction}
In recent years, there has been an increasing focus on neuromorphic or event-based vision due to its ability to excel under high dynamic range conditions, offer high temporal resolution, and consume less power than conventional frame-based vision sensors such as RGB cameras.
The event cameras, also known as dynamic vision sensors (DVS), mimic the behavior of biological retinas by continuously sampling incoming light and generating signals only when there is a change in light intensity. This results in an event data stream represented as a sequence of  $\langle x, y, p, t \rangle$  tuple, where $(x, y)$ denotes pixel position, $t$ represents time, and $p$ indicates polarity (positive or negative contrast)~\cite{gallego2020event}. Thus, event-based cameras represent a paradigm shift in environmental sensing and perception, capturing minute changes in local pixel-level light intensity and generating asynchronous event streams. This approach revolutionizes how autonomous and traffic ecosystems perceive and respond to their surroundings, enabling precise and efficient real-time processing of dynamic visual data~\cite{chen2020event, chakravarthi2023event, gallego2020event}.

While event-based sensing represents a novel area, research efforts have been limited in recent years to fully utilize the capabilities of event-based cameras for perception tasks. Notably, researchers have predominantly used event-based cameras like DAVIS346 by ini\uppercase{V}ation~\cite{iniVation} and Prophesee's IMX636 / EVK 4 HD~\cite{prophesee-evk4} to construct automotive datasets. Additionally, researchers have employed frame-to-event simulators such as ESIM~\cite{rebecq2018esim} and v2e~\cite{hu2021v2e} to generate synthetic event-based data. However, only \cite{hu2021v2e} converts RGB frames of an outdoor scene from MVSEC \cite{zhu2018multivehicle}. This highlights the significant scarcity of readily available synthetic event-based datasets in the field.
To bridge this gap and leverage the potential of synthetic data to generate diverse and high-quality vision data tailored for traffic monitoring, we present \textbf{\textit{SEVD}}  – a \textbf{S}ynthetic \textbf{E}vent-based \textbf{V}ision \textbf{D}ataset designed for autonomous driving and traffic monitoring tasks.

\textit{SEVD} is a multi-view dataset recorded using the CARLA~\cite{dosovitskiy2017carla} simulator, comprising ego and fixed perception data.
Ego perception data is captured through six DVS cameras providing a $360^\circ$ field-of-view (FoV) from a vehicle (as in Figure~\ref{fig01} (a)). Meanwhile, fixed perception data is recorded from four DVS sensors positioned at specific heights across four locations (as in Figure~\ref{fig01} (b)) at places like intersections, roundabouts, and underpasses, offering multiple views of the same site. Both ego and fixed perception data cover a wide range of environmental conditions. This includes diverse lighting scenarios such as noon, nighttime, and twilight, as well as various weather conditions like clear, cloudy, wet, soft-rainy, hard-rainy, and foggy. Additionally, the dataset spans various scenes, including urban, suburban, highway, and rural settings.

\begin{table*}[]
\centering
\resizebox{\textwidth}{!}{%
\begin{tabular}{c||c|c|c|cc|ccc|ccc|cccc|c|c|c|c}
\hline\hline
\multirow{2}{*}{} &
\multirow{2}{*}{\textbf{\begin{tabular}[c]{@{}c@{}}Dataset\end{tabular}}} &
\multirow{2}{*}{\textbf{Year}} &
\multirow{2}{*}{\textbf{\begin{tabular}[c]{@{}c@{}}Event\\ Camera\end{tabular}}} &
\multicolumn{2}{c|}{\textbf{Perspective}} &
\multicolumn{3}{c|}{\textbf{Objects}} &
\multicolumn{3}{c|}{\textbf{Lighting}} &
\multicolumn{4}{c|}{\textbf{Weather}} &
\multirow{2}{*}{\textbf{\begin{tabular}[c]{@{}c@{}}Multi-view\end{tabular}}} &
\multirow{2}{*}{\textbf{\begin{tabular}[c]{@{}c@{}}Other Sensors\end{tabular}}} &
\multirow{2}{*}{\textbf{Scene}} &
\multirow{2}{*}{\textbf{\begin{tabular}[c]{@{}c@{}}Other \\ Information\end{tabular}}} \\ \cline{5-16}
&
&
&
&
\multicolumn{1}{c|}{\textbf{Ego}} &
\textbf{Fixed} &
\multicolumn{1}{c|}{\textbf{VH}} &
\multicolumn{1}{c|}{\textbf{PED}} &
\textbf{MM} &
\multicolumn{1}{c|}{\textbf{DY}} &
\multicolumn{1}{c|}{\textbf{NT}} &
\textbf{TW} &
\multicolumn{1}{c|}{\textbf{CLR}} &
\multicolumn{1}{c|}{\textbf{CDY}} &
\multicolumn{1}{c|}{\textbf{RNY}} &
\textbf{FGY} &
&
&
&
\\ \hline\hline
\multirow{7}{*}{\rotatebox[origin=c]{90}{\centering \textbf{Real-world}}} & 
 {DDD17~\cite{binas2017ddd17}} &
2017 &
\begin{tabular}[c]{@{}c@{}}DAVIS 346B\\ $346\times260\,\text{px}$ 
\end{tabular} &
\multicolumn{1}{c|}{\checkmark} &
&
\multicolumn{1}{c|}{\checkmark} &
\multicolumn{1}{c|}{}
&
&
\multicolumn{1}{c|}{\checkmark} &
\multicolumn{1}{c|}{\checkmark} &
\checkmark &
\multicolumn{1}{c|}{\checkmark} &
\multicolumn{1}{c|}{} &
\multicolumn{1}{c|}{\checkmark} &
&
&
RGB &
\begin{tabular}[c]{@{}c@{}}Freeway, \\ Highway, City\end{tabular} &
\begin{tabular}[c]{@{}c@{}}$12\,\text{hr}$,\\ HDF5 format\end{tabular} \\ \cline{2-20} 
&
 {MVSEC~\cite{zhu2018multivehicle}} &
2018 &
\begin{tabular}[c]{@{}c@{}}DAVIS 346B\\ $346\times260\,\text{px}$ \end{tabular} &
\multicolumn{1}{c|}{\checkmark} &
&
\multicolumn{1}{c|}{\checkmark} &
\multicolumn{1}{c|}{} &
&
\multicolumn{1}{c|}{\checkmark} &
\multicolumn{1}{c|}{} &
&
\multicolumn{1}{c|}{\checkmark} &
\multicolumn{1}{c|}{} &
\multicolumn{1}{c|}{}
&
&
\begin{tabular}[c]{@{}c@{}}\checkmark \ (2 event \\ cameras used)\end{tabular} &
\begin{tabular}[c]{@{}c@{}}RGB, VLP-16 Lidar, \\ GPS\end{tabular} &
Indoor, Outdoor &
\begin{tabular}[c]{@{}c@{}}$\sim$ $1\,\text{hr}$,\\ rosbag format\end{tabular} \\ \cline{2-20} 
&
 {N-Cars~\cite{sironi2018hats}} &
2018 &
\begin{tabular}[c]{@{}c@{}}Prophesee Gen1\\  $304\times240\,\text{px}$ \end{tabular} &
\multicolumn{1}{c|}{\checkmark} &
&
\multicolumn{1}{c|}{\checkmark} &
\multicolumn{1}{c|}{} &
&
\multicolumn{1}{c|}{} &
\multicolumn{1}{c|}{} &
   &
  \multicolumn{1}{c|}{} &
  \multicolumn{1}{c|}{} &
  \multicolumn{1}{c|}{} &
   &
   &
  - &
  Urban &
  \begin{tabular}[c]{@{}c@{}}$1.2\,\text{hr}$\\ DAT format\end{tabular} \\ \cline{2-20} 
 &
   {DDD20~\cite{hu2020ddd20}} &
  2020 &
  \begin{tabular}[c]{@{}c@{}}DAVIS 346B\\ $346\times240\,\text{px}$\end{tabular} &
  \multicolumn{1}{c|}{\checkmark} &
   &
  \multicolumn{1}{c|}{\checkmark} &
  \multicolumn{1}{c|}{} &
   &
  \multicolumn{1}{c|}{\checkmark} &
  \multicolumn{1}{c|}{\checkmark} &
  \checkmark &
  \multicolumn{1}{c|}{\checkmark} &
  \multicolumn{1}{c|}{} &
  \multicolumn{1}{c|}{} &
   &
   &
  RGB &
  Highway, Urban &
  \begin{tabular}[c]{@{}c@{}}$39\,\text{hr}$,\\ HDF5 format\end{tabular} \\ \cline{2-20} 
 &
   {GEN1~\cite{de2020large}} &
  2020 &
  \begin{tabular}[c]{@{}c@{}}Prophesee Gen1\\ $346\times240\,\text{px}$\end{tabular} &
  \multicolumn{1}{c|}{\checkmark} &
   &
  \multicolumn{1}{c|}{\checkmark} &
  \multicolumn{1}{c|}{\checkmark} &
  \checkmark &
  \multicolumn{1}{c|}{\checkmark} &
  \multicolumn{1}{c|}{} &
   &
  \multicolumn{1}{c|}{\checkmark} &
  \multicolumn{1}{c|}{} &
  \multicolumn{1}{c|}{} &
   &
   &
   &
  \begin{tabular}[c]{@{}c@{}}City, Highway,\\ Suburban, Countryside\end{tabular} &
  \begin{tabular}[c]{@{}c@{}}$10\,\text{hr}$, DAT format\\ 255K bounding boxes\end{tabular} \\ \cline{2-20} 
 &
   {1Mpx~\cite{perot2020learning}} &
  2020 &
  \begin{tabular}[c]{@{}c@{}}Prophesee Gen2\\ $1280\times720\,\text{px}$\end{tabular} &
  \multicolumn{1}{c|}{\checkmark} &
   &
  \multicolumn{1}{c|}{\checkmark} &
  \multicolumn{1}{c|}{\checkmark} &
  \checkmark &
  \multicolumn{1}{c|}{\checkmark} &
  \multicolumn{1}{c|}{} &
   &
  \multicolumn{1}{c|}{\checkmark} &
  \multicolumn{1}{c|}{} &
  \multicolumn{1}{c|}{} &
   &
   &
  RGB &
  \begin{tabular}[c]{@{}c@{}}City, Highway,\\ Suburban, Countryside\end{tabular} &
  \begin{tabular}[c]{@{}c@{}}$14\,\text{hr}$, DAT format,\\ 25M bounding boxes\end{tabular} \\ \cline{2-20} 
 &
  {DSEC~\cite{gehrig2021dsec}} &
  2021 &
  \begin{tabular}[c]{@{}c@{}}Prophesee Gen3.1\\  $640\times480\,\text{px}$ \end{tabular} &
  \multicolumn{1}{c|}{\checkmark} &
   &
  \multicolumn{1}{c|}{\checkmark} &
  \multicolumn{1}{c|}{} &
   &
  \multicolumn{1}{c|}{\checkmark} &
  \multicolumn{1}{c|}{\checkmark} &
   &
  \multicolumn{1}{c|}{} &
  \multicolumn{1}{c|}{} &
  \multicolumn{1}{c|}{} &
   &
   &
  RGB, Lidar, GNSS &
  City &
  $\sim$$1\,\text{hr}$ \\ \hline
\multirow{2}{*}{\rotatebox[origin=c]{90}{\textbf{Synthetic}}} &
   {v2e~\cite{hu2021v2e}} &
  2021 &
  v2e simulator &
  \multicolumn{1}{c|}{\checkmark} &
   &
  \multicolumn{1}{c|}{\checkmark} &
  \multicolumn{1}{c|}{} &
   &
  \multicolumn{1}{c|}{\checkmark} &
  \multicolumn{1}{c|}{} &
   &
  \multicolumn{1}{c|}{} &
  \multicolumn{1}{c|}{} &
  \multicolumn{1}{c|}{} &
   &
   &
  - &
  Indoor, Outdoor
  &
  \begin{tabular}[c]{@{}c@{}}Frame-based from \\ MVSEC converted \\ to the event stream\end{tabular} \\ \cline{2-20} 
 &
  SEVD (Ours) &
  2024 &
  DVS &
  \multicolumn{1}{c|}{\checkmark} &
  \checkmark &
  \multicolumn{1}{c|}{\checkmark} &
  \multicolumn{1}{c|}{\checkmark} &
  \checkmark &
  \multicolumn{1}{c|}{\checkmark} &
  \multicolumn{1}{c|}{\checkmark} &
  \checkmark &
  \multicolumn{1}{c|}{\checkmark} &
  \multicolumn{1}{c|}{\checkmark} &
  \multicolumn{1}{c|}{\checkmark} &
  \checkmark &
  \begin{tabular}[c]{@{}c@{}}\checkmark\\ (6 Ego and 4 Fixed event cameras \\ used with a $360^\circ$ FoV)\end{tabular} &
  \begin{tabular}[c]{@{}c@{}}RGB, Depth, Optical, \\ Semantic, Instance, \\ GNSS, IMU\end{tabular} &
  \begin{tabular}[c]{@{}c@{}}Urban, Highway,\\ Suburban, Countryside, Intersections\end{tabular} &
  \begin{tabular}[c]{@{}c@{}}$58\,\text{hr}$, npz format,\\ $9\text{M}$  bounding boxes\end{tabular} 
   \\ \hline\hline
\end{tabular}%
}
\caption{A comprehensive overview of existing real and synthetic event-based automotive traffic perception datasets. (VH: Vehicle, PED: Pedestrian, MM: Micro-mobility, DY: Daytime, NT: Nighttime, TW: Twilight, CLR: Clear, CDY: Cloudy,  RNY: Rainy, FGY: Foggy)}
\label{table01}
\end{table*}

The event cameras are complemented by five different types of sensors, including RGB, depth, optical flow, semantic, and instance segmentation cameras, resulting in a diverse array of data as shown in Figure~\ref{fig02}. The dataset also includes GNSS and IMU data to provide additional context for driving scenarios. Annotations for various traffic participants, such as cars, trucks, vans, pedestrians, motorcycles, and bicycles, are provided in both $2\text{D}$  and $3\text{D}$ bounding boxes, following the COCO~\cite{lin2014microsoft}, Pascal VOC~\cite{everingham2015pascal}, and KITTI~\cite{Geiger2013IJRR} format respectively. \textit{SEVD} offers raw event streams $\langle x, y, p, t \rangle$ in \textit{.npz} format alongside their corresponding images. This dataset represents a significant advancement as the first-of-its-kind synthetic event-based data providing both ego and fixed perception, featuring a comprehensive range of annotations, extensive recording hours, and diverse driving conditions. We outline the key contributions of the presented work below:

\begin{enumerate}
\item A multi-view ($360^\circ$) synthetic event-based dataset comprising  $27\,\text{hr}$ of fixed and $31\,\text{hr}$ of ego perception data, with over $9\text{M}$  bounding boxes, recorded across diverse conditions and varying parameters.
\item Establishing benchmark baselines for $2\text{D}$ detection using state-of-the-art event-based and frame-based architectures on \textit{SEVD} across different driving and fixed traffic monitoring scenarios to assess the efficacy of the dataset.
\item A quantitative and qualitative evaluation of synthetic event-based detector's generalization capabilities on real-world data.
\end{enumerate}

The rest of this paper is structured as follows. Section~\ref{sec:relStudy} briefly summarizes the existing event-based dataset contributions. Section~\ref{sec:SEVD_overview} presents the \textit{SEVD} dataset, detailing the generation framework and annotation process. In Section~\ref{sec:experiments}, we provide baselines and explore the generalization capabilities of synthetic event-based detectors to real-world data.

\section{Related study}\label{sec:relStudy}
With recent advancements in neuromorphic vision, researchers have curated event-based datasets across various domains~\cite{mueggler2017event, li2020event, boretti2023pedro, amir2017low, Bolten_2021_CVPR}. This section provides an overview of event-based automotive traffic perception datasets, focusing on their availability in real or synthetic environments, as summarized in Table~\ref{table01}.

\subsection{Event-based real automotive datasets}
Real event-based automotive datasets are crucial as they offer valuable insights for advancing autonomous driving research. The DDD17~\cite{binas2017ddd17} is the first open dataset offering driving recordings, annotated data for end-to-end learning approaches, and sensor fusion techniques. Meanwhile, the MVSEC~\cite{zhu2018multivehicle} dataset addresses the scarcity of labeled data for $3\text{D}$ perception tasks. It offers synchronized stereo pair event-based data captured in diverse scenarios, enabling the development and testing of algorithms for tasks like feature tracking, visual odometry, and stereo depth estimation.

Additionally, N-Cars~\cite{sironi2018hats} serves a large dataset for object classification, showcasing improved classification performance in real-time computation for applications like autonomous vehicles and UAV vision. 
Furthermore, the DDD20~\cite{hu2020ddd20} expands DDD17~\cite{binas2017ddd17} with an additional $39\,\text{hr}$ of data, making it the longest event camera end-to-end driving dataset. 
The GEN1~\cite{de2020large}  automotive dataset offers over $39\,\text{hr}$ of recordings captured with a $304\times240\,\text{px}$ GEN1 sensor. It provides manual annotations for cars and pedestrians, contributing to the advancement of event-based vision tasks such as object detection and classification. The 1 Megapixel Automotive Dataset~\cite{perot2020learning} is a large-scale and high-resolution dataset containing over $14\,\text{hr}$ of recordings with $25\text{M}$ bounding boxes of cars, pedestrians, and two-wheelers labeled at high frequency in automotive scenarios.

Lastly, DSEC~\cite{gehrig2021dsec}, a unique dataset designed for driving scenarios in challenging illumination conditions. It is the first large-scale stereo dataset with event cameras containing $53$ sequences collected in various illumination conditions. This dataset provides ground truth disparity for developing and evaluating event-based stereo algorithms, advancing autonomous driving research.

\begin{figure*}[ht]
  \centering
\includegraphics[width=0.92\linewidth]{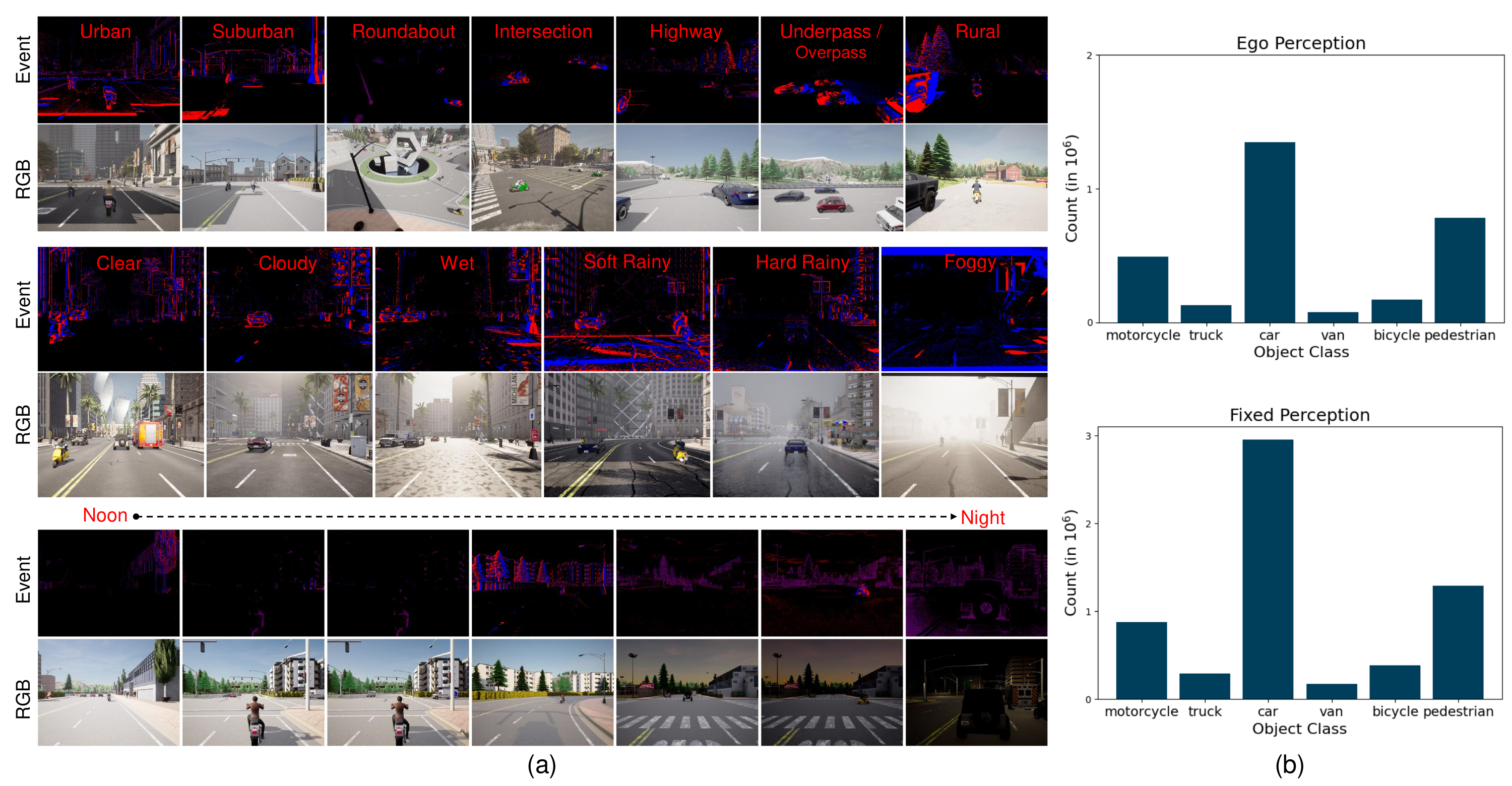}
\caption{(a) Visual Representation (event and RGB) of \textit{SEVD} Dataset Features: The scene diversity (top row) considered during data generation from the urban to the rural, weather variability (middle row) captured in the dataset, ranging from clear skies to foggy scenarios, and the dynamic conditions (bottom row)  showcasing sequences with continuously shifting parameters, mirroring real-world driving scenarios.  (b) Dataset Distribution: shows the distribution of instances for each class in both ego and fixed perception scenarios.}
\label{fig03}
\end{figure*}

\subsection{Event-based synthetic automotive datasets}
In the synthetic event data generation space, prominent simulators include the RPG DAVIS~\cite{mueggler2017event}, ESIM~\cite{rebecq2018esim}, v2e~\cite{hu2021v2e}, blinkSim~\cite{li2023blinkflow}, and the CARLA Simulator~\cite{dosovitskiy2017carla}. The v2e simulator stands out for its ability to generate realistic synthetic DVS events from intensity frames and ensures realism. Notably, v2e utilized the MVSEC~\cite{zhu2018multivehicle} dataset to convert RGB to event frames of an outdoor scenario during nighttime, providing vehicle annotations. However, there are no extensive synthetic event-based automotive traffic perception datasets to the best of our knowledge. Unlike the aforementioned simulators, which primarily transform existing frame-based datasets into their event counterparts, CARLA generates event data by uniformly sampling between two consecutive synchronous frames in dynamic traffic conditions. Interestingly, despite its potential, the DVS capabilities of the CARLA simulator remain largely unexplored. 

\section{SEVD - Dataset overview}
\label{sec:SEVD_overview}
This section provides an overview of the sensor suite and multi-camera configuration employed for capturing both ego and fixed perception data. We discuss the data generation pipeline implemented within the CARLA simulator environment, describe the labeling protocol, and a breakdown of the generated data. 

\subsection{Sensor setup}
The sensor suite comprises a strategically positioned array of sensors of each type (event, RGB, depth, optical flow, semantic, and instance), tailored for both ego and fixed scenarios. In ego scenarios, the cameras offer coverage from front to rear, including front-right, front-left, rear-right, and rear-left perspectives, each with overlapping FoV providing a comprehensive $360^\circ$ view. Notably, the rear camera features a wider $110^\circ$ FoV, while the others have a $70^\circ$ FoV,  following the approach used in nuScenes~\cite{caesar2020nuscenes
}. For fixed-perception scenarios, four cameras, each offering a $90^\circ$ FoV, are used. This strategic setup enhances our data collection strategy, contributing to a diverse range of data across various conditions. 

The DVS camera within the CARLA environment is configured with a dynamic range of $140\,\text{dB}$, a temporal resolution in the order of microseconds, and a resolution of $1280\times960\,\text{px}$. It generates a continuous stream of events represented by $\langle x, y, p, t \rangle$, where an event is triggered at pixel coordinates $(x, y)$ and timestamp $t$ when the logarithmic intensity change $L$ exceeds a predefined constant threshold $C$, as defined by the equation below \cite{carla_dvs}:
\begin{equation}
    L(x, y, t) - L(x, y, t - \delta t) = \text{p} \times C
\end{equation}

\begin{math}
    t - \delta t
\end{math} is the time when the last event at that pixel was triggered and $p$ represents the polarity of the event: \( p = +1 \) indicates an increment in brightness, and \( p = -1 \) indicates a decrement, with $C = 0.3$.

\begin{table*}[]
\centering
\resizebox{0.95\linewidth}{!}{%
\begin{tabular}{c|c|cccc|cccc|cccc}
\hline\hline
\multirow{2}{*}{\textbf{\begin{tabular}[c]{@{}c@{}}Map\\ Type\end{tabular}}} &
  \multirow{2}{*}{\textbf{\begin{tabular}[c]{@{}c@{}}Lighting \\ Conditions\end{tabular}}} &
  \multicolumn{4}{c|}{\textbf{RED (EVENT)}} &
  \multicolumn{4}{c||}{\textbf{RVT (EVENT)}} &
  \multicolumn{4}{c}{\textbf{YOLOv8 (RGB)}} 
  \\ \cline{3-14} 
 &
   &
  \multicolumn{1}{c|}{\textbf{Car}} &
  \multicolumn{1}{c|}{\textbf{Pedestrian}} &
  \multicolumn{1}{c|}{\textbf{Motorcycle}} &
  \textbf{All Classes} &
  \multicolumn{1}{c|}{\textbf{Car}} &
  \multicolumn{1}{c|}{\textbf{Pedestrian}} &
  \multicolumn{1}{c|}{\textbf{Motorcycle}} &
  \multicolumn{1}{c||}{\textbf{All Classes}} &
  \multicolumn{1}{c|}{\textbf{Car}} &
  \multicolumn{1}{c|}{\textbf{Pedestrian}} &
  \multicolumn{1}{c|}{\textbf{Motorcycle}} &
  \multicolumn{1}{c}{\textbf{All Classes}}
  \\ \hline \hline  
\textbf{Intersection} &
  \multirow{4}{*}{\textbf{Noon}} &
  \multicolumn{1}{c|}{0.530 } &
  \multicolumn{1}{c|}{ 0.386} &
  \multicolumn{1}{c|}{ 0.415} &
  \multicolumn{1}{c|}{ 0.387} &
  \multicolumn{1}{c|}{ 0.537} &
  \multicolumn{1}{c|}{0.810} &
  \multicolumn{1}{c|}{ 0.551} &
  \multicolumn{1}{c||}{ 0.552} &
  \multicolumn{1}{c|}{0.801} &
  \multicolumn{1}{c|}{0.439} &
  \multicolumn{1}{c|}{0.697} &
  \multicolumn{1}{c}{0.659} 
    \\ \cline{1-1} \cline{3-14} 
\textbf{Roundabout} &
   &
  \multicolumn{1}{c|}{ 0.865} &
  \multicolumn{1}{c|}{0.439} &
  \multicolumn{1}{c|}{0.770} & 
   \multicolumn{1}{c|}{0.728} &
  \multicolumn{1}{c|}{0.841 } &
  \multicolumn{1}{c|}{- } &
  \multicolumn{1}{c|}{ 0.846} & 
   \multicolumn{1}{c||}{0.854} &
  \multicolumn{1}{c|}{0.987} &
  \multicolumn{1}{c|}{ 0.475} &
  \multicolumn{1}{c|}{ 0.957 } &
  \multicolumn{1}{c}{0.886} 
    \\ \cline{1-1} \cline{3-14} 
\textbf{Underpass / Overpass} &
   &
  \multicolumn{1}{c|}{ 0.779} &
  \multicolumn{1}{c|}{ -} &
  \multicolumn{1}{c|}{- } & 0.756
    &
  \multicolumn{1}{c|}{0.955 } &
  \multicolumn{1}{c|}{ -} &
  \multicolumn{1}{c|}{ -} & 
  \multicolumn{1}{c||}{0.944}
    &
 \multicolumn{1}{c|}{ 0.823} &
  \multicolumn{1}{c|}{ -} &
  \multicolumn{1}{c|}{-} &
  \multicolumn{1}{c}{0.828} 
  
    \\ \cline{1-1} \cline{3-14} 
\textbf{All Maps} &
   &
  \multicolumn{1}{c|}{ 0.680} &
  \multicolumn{1}{c|}{ 0.398} &
  \multicolumn{1}{c|}{ 0.575} &0.520
    &
  \multicolumn{1}{c|}{0.679 } &
  \multicolumn{1}{c|}{0.809} &
  \multicolumn{1}{c|}{0.694 } & 
  \multicolumn{1}{c||}{0.683}
    &
  \multicolumn{1}{c|}{ 0.838 } &
  \multicolumn{1}{c|}{0.441} &
  \multicolumn{1}{c|}{0.778} &
  \multicolumn{1}{c}{0.717} 
    \\ \hline \hline 
\textbf{Intersection} &
  \multirow{4}{*}{\textbf{Night}} &
  \multicolumn{1}{c|}{0.451 } &
  \multicolumn{1}{c|}{0.331 } &
  \multicolumn{1}{c|}{ 0.144} &0.310
    & 
  \multicolumn{1}{c|}{0.469 } &
  \multicolumn{1}{c|}{0.650} &
  \multicolumn{1}{c|}{0.228} &
  \multicolumn{1}{c||}{0.477}
    &
  \multicolumn{1}{c|}{0.737} &
  \multicolumn{1}{c|}{0.328} &
  \multicolumn{1}{c|}{0.613} &
  \multicolumn{1}{c}{0.619} 
    \\ \cline{1-1} \cline{3-14} 
\textbf{Roundabout} &
   &
  \multicolumn{1}{c|}{ 0.488} &
  \multicolumn{1}{c|}{ 0.235} &
  \multicolumn{1}{c|}{ 0.472} & 0.548
    &
  \multicolumn{1}{c|}{0.547 } &
  \multicolumn{1}{c|}{ -} &
  \multicolumn{1}{c|}{0.566 } &
  \multicolumn{1}{c||}{0.626}
    &
  \multicolumn{1}{c|}{0.747} &
  \multicolumn{1}{c|}{0.220} &
  \multicolumn{1}{c|}{0.797} &
  \multicolumn{1}{c}{0.738} 
    \\ \cline{1-1} \cline{3-14} 
\textbf{Underpass / Overpass} &
   &
  \multicolumn{1}{c|}{ 0.693} &
  \multicolumn{1}{c|}{- } &
  \multicolumn{1}{c|}{ -} & 0.473
    &
  \multicolumn{1}{c|}{0.357 } &
  \multicolumn{1}{c|}{-} &
  \multicolumn{1}{c|}{-} &
  \multicolumn{1}{c||}{0.315}
    &
  \multicolumn{1}{c|}{0.774} &
  \multicolumn{1}{c|}{-} &
  \multicolumn{1}{c|}{- } &
  \multicolumn{1}{c}{0.769} 
    \\ \cline{1-1} \cline{3-14} 
\textbf{All Maps} &
   &
  \multicolumn{1}{c|}{0.467 } &
  \multicolumn{1}{c|}{0.313 } &
  \multicolumn{1}{c|}{0.229 } & 0.398
    &
  \multicolumn{1}{c|}{0.477 } &
  \multicolumn{1}{c|}{0.644 } &
  \multicolumn{1}{c|}{0.328 } & 
  \multicolumn{1}{c||}{0.496}
    &
  \multicolumn{1}{c|}{0.736 } &
  \multicolumn{1}{c|}{ 0.303} &
  \multicolumn{1}{c|}{ 0.671} &
  \multicolumn{1}{c}{0.667 } 
    \\ \hline \hline 
\textbf{Intersection} &
  \multirow{4}{*}{\textbf{Twilight}} &
  \multicolumn{1}{c|}{ 0.479} &
  \multicolumn{1}{c|}{ 0.323} &
  \multicolumn{1}{c|}{ 0.398} &0.370
    &
  \multicolumn{1}{c|}{0.574 } &
  \multicolumn{1}{c|}{0.720} &
  \multicolumn{1}{c|}{ 0.601} &
  \multicolumn{1}{c||}{0.600}
    &
  \multicolumn{1}{c|}{0.658} &
  \multicolumn{1}{c|}{ 0.334} &
  \multicolumn{1}{c|}{0.610} &
  \multicolumn{1}{c}{0.577} 
    \\ \cline{1-1} \cline{3-14} 
\textbf{Roundabout} &
  \multicolumn{1}{l|}{} &
  \multicolumn{1}{c|}{ 0.902} &
  \multicolumn{1}{c|}{ 0.454} &
  \multicolumn{1}{c|}{ 0.742} & 0.760
    &
  \multicolumn{1}{c|}{ 0.884} &
  \multicolumn{1}{c|}{ -} &
  \multicolumn{1}{c|}{ 0.888} &
  \multicolumn{1}{c||}{0.896}
    &
  \multicolumn{1}{c|}{0.987} &
  \multicolumn{1}{c|}{0.583} &
  \multicolumn{1}{c|}{0.938} &
  \multicolumn{1}{c}{0.903} 
    \\ \cline{1-1} \cline{3-14} 
\textbf{Underpass / Overpass} &
  \multicolumn{1}{l|}{} &
  \multicolumn{1}{c|}{0.777} &
  \multicolumn{1}{c|}{ -} &
  \multicolumn{1}{c|}{-} & 0.669
    &
  \multicolumn{1}{c|}{0.961 } &
  \multicolumn{1}{c|}{- } &
  \multicolumn{1}{c|}{ -} &
  \multicolumn{1}{c||}{0.942}
    &
  \multicolumn{1}{c|}{0.840} &
  \multicolumn{1}{c|}{- } &
  \multicolumn{1}{c|}{ -} &
  \multicolumn{1}{c}{0.793} 
    \\ \cline{1-1} \cline{3-14} 
\textbf{All Maps} &
  \multicolumn{1}{l|}{} &
  \multicolumn{1}{c|}{ 0.571} &
  \multicolumn{1}{c|}{ 0.346} &
  \multicolumn{1}{c|}{0.439 } & 0.474
    &
  \multicolumn{1}{c|}{ 0.650} &
  \multicolumn{1}{c|}{0.720} &
  \multicolumn{1}{c|}{ 0.630} & 
  \multicolumn{1}{c||}{0.670}
    &
  \multicolumn{1}{c|}{0.708 } &
  \multicolumn{1}{c|}{0.362 } &
  \multicolumn{1}{c|}{ 0.648} &
  \multicolumn{1}{c}{0.647 } 
    \\ \hline \hline 
\end{tabular}%
}
\caption{Fixed perception Baseline Evaluation: Results of tensor-based methods RVT and RED, along with the frame-based approach YOLOv8, across different map types (Intersections, Roundabout, Underpass, and Overpass) in fixed perception scenarios. RVT and RED evaluations are conducted on event data, while YOLOv8 is evaluated on RGB data.}
\label{tab:fixed_baselines}
\end{table*}

Furthermore, the RGB, depth, optical flow, semantic, and instance segmentation cameras are all configured to operate at a resolution of $1280\times960\,\text{px}$.
Depth information is encoded into grayscale images with floating-point values between [$0$,$1$] and with a $1\,\text{mm}$ resolution. The optical flow, crucial for motion-centric algorithms, is provided in UV map format. For each frame, we provide panoptic segmentation labels, encompassing both instance and semantic segmentation, across the $23$ classes defined in the Cityscapes \cite{cordts2015cityscapes} annotation scheme.  
Additionally, GNSS and IMU sensors are incorporated to track the position and orientation of the vehicle precisely in ego perception scenarios.

\subsection{Data generation and annotation}
We leverage the CARLA simulator's robust data generation pipeline to record traffic data across diverse scenes (see Figure~\ref{fig03} (a), top row), illumination, weather conditions (see Figure~\ref{fig03} (a), middle row)), and varying traffic densities. Each data generation iteration initializes the simulation with custom settings and generates traffic. The data sequences are generated under discrete environmental weather conditions, with each set of sequences collected using different domain parameters and initial states. Additionally, we generate sequences with continuously shifting conditions (see Figure~\ref{fig03} (a), bottom row), where a shift is generated by interpolating the state between the initial and final frame.
To ensure precise synchronization among sensors, the simulation operates in synchronous mode with a time-step of $0.1$ seconds, corresponding to $10$ frames per second (fps). Following the simulation phase, we employ CARLA's bounding box API to generate annotations in both $2\text{D}$ and $3\text{D}$ bounding box formats, such as COCO, Pascal VOC, and KITTI. Additionally, we offer annotations for optical flow and dense depth. We provide code for custom data generation at~\url{https://github.com/eventbasedvision/SEVD}  

\subsection{Recording and statistics}
\textit{SEVD} offers a diverse range of recordings featuring various combinations of scenes, weather, and lighting conditions. For instance, in an ego perception scenario, recordings may feature an ego vehicle navigating through a suburban environment under a continuous domain shift, transitioning from noon to night in clear weather conditions as depicted in Figure~\ref{fig03} (a) bottom row. Another example of a fixed perception may feature a twilight setting with a soft rain over a four-way intersection. Each recording spans durations of $2$  to $30\,\text{min}$. We provide a total of $27\,\text{hr}$ of fixed and $31\,\text{hr}$ of ego perception event data collectively. Similarly, we offer an equal volume of data from other sensor types, resulting in a cumulative $162\,\text{hr}$ of fixed and $186\,\text{hr}$ of ego perception data. 

\textit{SEVD} comprises extensive annotations, including  $2\text{D}$ and $3\text{D}$ bounding boxes for six categories (car, truck, bus, bicycle, motorcycle, and pedestrian)  of traffic participants, totaling approximately $9\text{M}$ bounding boxes, with cars being the most prevalent category as illustrated in Figure~\ref{fig03}~(b).
To facilitate model training, the dataset is segmented into subsets containing 70\% train, 15\% validation, and 15\% test data, ensuring proportional representation across various combinations.

\section{Experiments}
\label{sec:experiments}
\textit{SEVD} is a comprehensive dataset, offering both ego and fixed perception multi-view multi-sensor data, particularly emphasizing event-based vision for autonomous driving and traffic monitoring. In our experiments, we focus on modeling event data for the $2\text{D}$ detection task across various scenes, weather conditions, and lighting scenarios.
To establish baselines, we train state-of-the-art event-based detectors, including Recurrent Vision Transformers (RVT)~\cite{gehrig2023recurrent}, Recurrent Event-camera Detector (RED)~\cite{perot2020learning}. Additionally, we evaluate frame-based (RGB) data using the You Only Look Once (YOLOv8)~\cite{Jocher_Ultralytics_YOLO_2023} detector. While our primary focus lies in event data, \textit{SEVD} also provides other modality data to support broader research endeavors. Additionally, we conduct qualitative and quantitative evaluations for event-based detectors using real-world event data, offering insights into their generalization performance in dynamic scenarios.
For all experiments, the evaluation metrics are based on Mean Average Precision (mAP) at a 50\% Intersection over Union (IoU) threshold. 

\begin{table*}[t]
\centering
\resizebox{0.95\textwidth}{!}{%
\begin{tabular}{c|c|cccc|cccc|cccc||}
\hline \hline
\multirow{2}{*}{\textbf{\begin{tabular}[c]{@{}c@{}}Map\\ Type\end{tabular}}} &
  \multirow{2}{*}{\textbf{\begin{tabular}[c]{@{}c@{}}Lighting \\ Conditions\end{tabular}}} &
  \multicolumn{4}{c|}{\textbf{RED (EVENT)}} &
  \multicolumn{4}{c||}{\textbf{RVT (EVENT)}} &
  \multicolumn{4}{c}{\textbf{YOLOv8 (RGB)}} 
 \\ \cline{3-14} 
 &
   &
  \multicolumn{1}{c|}{\textbf{Car}} &
  \multicolumn{1}{c|}{\textbf{Pedestrian}} &
  \multicolumn{1}{c|}{\textbf{Motorcycle}} &
  \multicolumn{1}{c|}{\textbf{All Classes}} &
  \multicolumn{1}{c|}{\textbf{Car}} &
  \multicolumn{1}{c|}{\textbf{Pedestrian}} &
  \multicolumn{1}{c|}{\textbf{Motorcycle}} &
  \multicolumn{1}{c||}{\textbf{All Classes}} &
  \multicolumn{1}{c|}{\textbf{Car}} &
  \multicolumn{1}{c|}{\textbf{Pedestrian}} &
  \multicolumn{1}{c|}{\textbf{Motorcycle}} &
  \multicolumn{1}{c}{\textbf{All Classes}}  \\ \hline \hline
\textbf{Urban} &
  \multirow{5}{*}{\textbf{Noon}} &
  \multicolumn{1}{c|}{0.524} &
  \multicolumn{1}{c|}{0.124} &
  \multicolumn{1}{c|}{0.095} &0.167
   &
  \multicolumn{1}{c|}{0.598} &
  \multicolumn{1}{c|}{0.149} &
  \multicolumn{1}{c|}{0.268} & 
  \multicolumn{1}{c||}{0.444}&
   
     \multicolumn{1}{c|}{0.961} &
  \multicolumn{1}{c|}{0.63} &
    \multicolumn{1}{c|}{0.806} &
  \multicolumn{1}{c}{0.881}
   \\ \cline{1-1} \cline{3-14} 
\textbf{Suburban} &
   &
  \multicolumn{1}{c|}{0.375} &
  \multicolumn{1}{c|}{0.057} &
  \multicolumn{1}{c|}{0.430} &0.239
   &
  \multicolumn{1}{c|}{0.523} &
  \multicolumn{1}{c|}{0.338} &
  \multicolumn{1}{c|}{0.714} & 
  \multicolumn{1}{c||}{0.618}
   &
  \multicolumn{1}{c|}{0.887} &
  \multicolumn{1}{c|}{0.376} &
  \multicolumn{1}{c|}{0.815} &
  \multicolumn{1}{c}{0.744} 
   \\ \cline{1-1} \cline{3-14} 
\textbf{Rural} &
   &
  \multicolumn{1}{c|}{0.418} &
  \multicolumn{1}{c|}{0.068} &
  \multicolumn{1}{c|}{0.226} &0.143
   &
  \multicolumn{1}{c|}{0.488} &
  \multicolumn{1}{c|}{0.473} &
  \multicolumn{1}{c|}{0.396} & 
  \multicolumn{1}{c||}{0.471}
   & 
  \multicolumn{1}{c|}{0.878} &
  \multicolumn{1}{c|}{0.46} &
  \multicolumn{1}{c|}{0.584} &
  \multicolumn{1}{c}{0.552} 
  \\ \cline{1-1} \cline{3-14} 
\textbf{Highway} &

   &
  \multicolumn{1}{c|}{0.248} &
  \multicolumn{1}{c|}{-} &
  \multicolumn{1}{c|}{-} &0.101
   &
  \multicolumn{1}{c|}{0.500} &
  \multicolumn{1}{c|}{-} &
  \multicolumn{1}{c|}{-} & 
  \multicolumn{1}{c||}{0.485}
   &
  \multicolumn{1}{c|}{0.881} &
  \multicolumn{1}{c|}{-} &
  \multicolumn{1}{c|}{-} &
  \multicolumn{1}{c}{0.870}

  \\ \cline{1-1} \cline{3-14} 
\textbf{All Maps} &
   &
  \multicolumn{1}{c|}{0.348} &
  \multicolumn{1}{c|}{0.098} &
  \multicolumn{1}{c|}{0.270} &0.159
   &
  \multicolumn{1}{c|}{0.514} &
  \multicolumn{1}{c|}{0.243} &
  \multicolumn{1}{c|}{0.565} & 
  \multicolumn{1}{c||}{0.515}
  &
  \multicolumn{1}{c|}{0.892} &
  \multicolumn{1}{c|}{0.536} &
  \multicolumn{1}{c|}{0.744} &
  \multicolumn{1}{c}{0.753} 
  
   \\ \hline \hline
   
\textbf{Urban} &
  \multirow{5}{*}{\textbf{Night}} &
  \multicolumn{1}{c|}{0.085} &
  \multicolumn{1}{c|}{0.080} &
  \multicolumn{1}{c|}{0.053} & 0.048
   &
  \multicolumn{1}{c|}{0.175} &
  \multicolumn{1}{c|}{0.228} &
  \multicolumn{1}{c|}{0.290} & 
  \multicolumn{1}{c||}{0.223}
   &
  \multicolumn{1}{c|}{0.607} &
  \multicolumn{1}{c|}{0.642} &
  \multicolumn{1}{c|}{0.478} &
  \multicolumn{1}{c}{0.668}

  \\ \cline{1-1} \cline{3-14} 
  
\textbf{Suburban} &
   &
  \multicolumn{1}{c|}{0.409} &
  \multicolumn{1}{c|}{0.159} &
  \multicolumn{1}{c|}{0.212} &0.202
   &
  \multicolumn{1}{c|}{0.735} &
  \multicolumn{1}{c|}{0.054} &
  \multicolumn{1}{c|}{0.439} & 
  \multicolumn{1}{c||}{0.534}
   &
  \multicolumn{1}{c|}{0.836} &
  \multicolumn{1}{c|}{0.641} &
  \multicolumn{1}{c|}{0.847} &
  \multicolumn{1}{c}{0.825} 
  \\ \cline{1-1} \cline{3-14} 
  
\textbf{Rural} &
   &
  \multicolumn{1}{c|}{0.115} &
  \multicolumn{1}{c|}{0.076} &
  \multicolumn{1}{c|}{0.030} &0.044
   &
  \multicolumn{1}{c|}{0.190} &
  \multicolumn{1}{c|}{0.129} &
  \multicolumn{1}{c|}{0.010} & 
  \multicolumn{1}{c||}{0.145}
   &
  \multicolumn{1}{c|}{0.727} &
  \multicolumn{1}{c|}{0.353} &
  \multicolumn{1}{c|}{0.396} &
  \multicolumn{1}{c}{0.493} 
  \\ \cline{1-1} \cline{3-14} 
  
\textbf{Highway} &
   &
  \multicolumn{1}{c|}{0.046} &
  \multicolumn{1}{c|}{-} &
  \multicolumn{1}{c|}{-} & 0.034
   &
  \multicolumn{1}{c|}{0.158} &
  \multicolumn{1}{c|}{-} &
  \multicolumn{1}{c|}{-} & 
  \multicolumn{1}{c||}{0.127}
   &
  \multicolumn{1}{c|}{0.559} &
  \multicolumn{1}{c|}{-} &
  \multicolumn{1}{c|}{-} &
  \multicolumn{1}{c}{0.456} 
   \\ \cline{1-1} \cline{3-14} 
   
\textbf{All Maps} &
   &
  \multicolumn{1}{c|}{0.166} &
  \multicolumn{1}{c|}{0.074} &
  \multicolumn{1}{c|}{0.088} &0.085
   &
  \multicolumn{1}{c|}{0.284} &
  \multicolumn{1}{c|}{0.175} &
  \multicolumn{1}{c|}{0.229} & 
  \multicolumn{1}{c||}{0.260}
   &
  \multicolumn{1}{c|}{ 0.701} &
  \multicolumn{1}{c|}{ 0.472} &
  \multicolumn{1}{c|}{0.525} &
  \multicolumn{1}{c}{0.649} 
  
   \\ \hline \hline
\textbf{Urban} &
  \multirow{5}{*}{\textbf{Twilight}} &
  \multicolumn{1}{c|}{0.307} &
  \multicolumn{1}{c|}{0.121} &
  \multicolumn{1}{c|}{0.330} &0.306
   &
  \multicolumn{1}{c|}{0.499} &
  \multicolumn{1}{c|}{0.073} &
  \multicolumn{1}{c|}{0.637} & 
  \multicolumn{1}{c||}{0.573}
   &
  \multicolumn{1}{c|}{0.909} &
  \multicolumn{1}{c|}{0.46} &
  \multicolumn{1}{c|}{0.765} &
  \multicolumn{1}{c}{0.721} 
   \\ \cline{1-1} \cline{3-14} 
\textbf{Suburban} &
   &
  \multicolumn{1}{c|}{0.323} &
  \multicolumn{1}{c|}{0.161} &
  \multicolumn{1}{c|}{0.136} &0.164
   &
  \multicolumn{1}{c|}{0.390} &
  \multicolumn{1}{c|}{0.279} &
  \multicolumn{1}{c|}{0.387} & 
  \multicolumn{1}{c||}{0.294}
   &
  \multicolumn{1}{c|}{0.772} &
  \multicolumn{1}{c|}{0.671} &
  \multicolumn{1}{c|}{0.764} &
  \multicolumn{1}{c}{0.713} 
  \\ \cline{1-1} \cline{3-14} 
\textbf{Rural} &
   &
  \multicolumn{1}{c|}{0.256} &
  \multicolumn{1}{c|}{0.151} &
  \multicolumn{1}{c|}{0.289} &0.186
   &
  \multicolumn{1}{c|}{0.265} &
  \multicolumn{1}{c|}{0.366} &
  \multicolumn{1}{c|}{0.258} & 
  \multicolumn{1}{c||}{0.311}
   &
  \multicolumn{1}{c|}{0.775} &
  \multicolumn{1}{c|}{0.616} &
  \multicolumn{1}{c|}{0.544} &
  \multicolumn{1}{c}{0.597} 
   \\ \cline{1-1} \cline{3-14} 
\textbf{Highway} &
   &
  \multicolumn{1}{c|}{0.502} &
  \multicolumn{1}{c|}{-} &
  \multicolumn{1}{c|}{-} & 0.267
   &
  \multicolumn{1}{c|}{0.566} &
  \multicolumn{1}{c|}{-} &
  \multicolumn{1}{c|}{-} & 
  \multicolumn{1}{c||}{0.590}
   &
  \multicolumn{1}{c|}{0.878} &
  \multicolumn{1}{c|}{-} &
  \multicolumn{1}{c|}{-} &
  \multicolumn{1}{c}{0.924} 
  
  \\ \cline{1-1} \cline{3-14} 
\textbf{All Maps} &
   &
  \multicolumn{1}{c|}{0.381} &
  \multicolumn{1}{c|}{0.128} &
  \multicolumn{1}{c|}{0.287} &0.211
   &
  \multicolumn{1}{c|}{0.485} &
  \multicolumn{1}{c|}{0.249} &
  \multicolumn{1}{c|}{0.569} & 
  \multicolumn{1}{c||}{0.485}
   &
  \multicolumn{1}{c|}{0.854} &
  \multicolumn{1}{c|}{0.559} &
  \multicolumn{1}{c|}{0.712} &
  \multicolumn{1}{c}{0.720} 

   \\ \hline \hline
\end{tabular}%
}
\caption{Ego Perception Baseline Evaluation: Results of tensor-based methods RVT and RED, along with the frame-based approach YOLOv8, across different map types (Urban, Suburban, and Rural) in ego-driving scenarios. RVT and RED evaluations are conducted on event data, while YOLOv8 is evaluated on RGB data captured from the front view of the ego-vehicle.}
\label{tab:ego_baselines}
\end{table*} 
\subsection{Baseline evaluation}

To evaluate detector performance on \textit{SEVD}, we train them using $10\,\text{hr}$ of data and assess them on separate sets comprising $2.5\,\text{hr}$ for validation and $2.5\,\text{hr}$ for testing. RVT and RED were trained from scratch over 4 days using an A100 GPU, while YOLOv8 underwent training for $3.5\,\text{days}$ on an A2000 GPU. We conducted distinct evaluations to discern how each model operates across various scenes, weather conditions, and lighting scenarios, encompassing both ego and fixed perception settings. This approach enables us to gain insights into the models' adaptability to diverse environmental contexts.
\begin{figure}[t]
  \centering
  \includegraphics[width=1\linewidth]{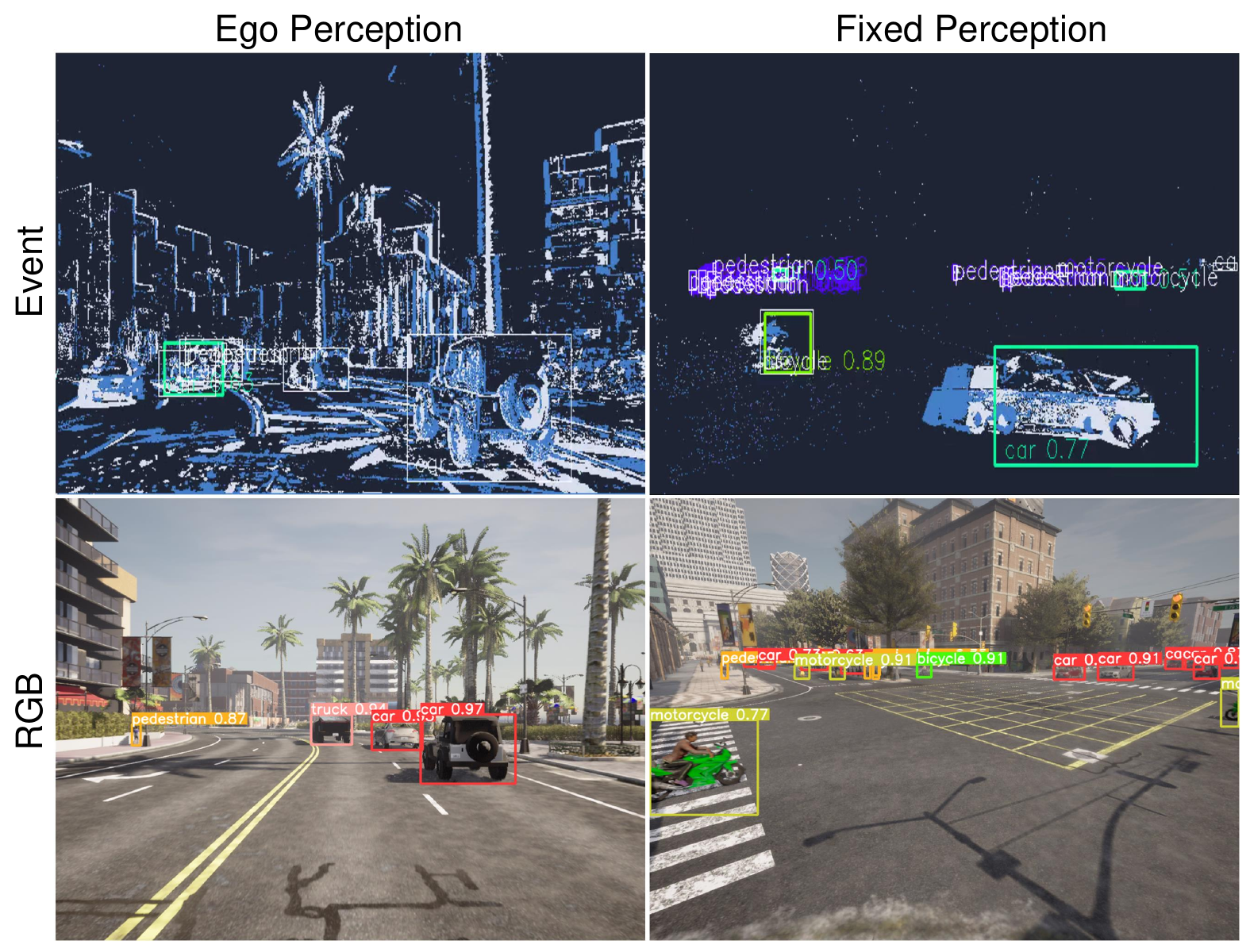}
   \caption{Qualitative Results: Showcasing event-based and frame-based detection of different classes of objects in ego (left column) and fixed  (right column) perception scenarios.}
   \label{fig04}
\end{figure}
Several key observations emerged from the evaluation of fixed and ego perception using tensor-based and frame-based models, as depicted in Table~\ref{tab:fixed_baselines} and Table~\ref{tab:ego_baselines}.

For the fixed perception scenario, we utilize all four views from the sensor setup (refer to Figure~\ref{fig02}~(b)).  Similar performance is noted among RED and RVT for car class detection, with RVT outperforming RED for pedestrian and motorcycle classes across all scenes and lighting conditions. Overall, RVT exhibits better detection performance across all classes and lighting conditions compared to RED. For frame-based detection, YOLOv8 demonstrates competitive performance in detecting cars across different lighting conditions but lags in pedestrian detection.
For ego perception scenarios, we utilize front view data of the ego vehicle. RVT outperforms RED in terms of all class detection across all scenes and lighting. Consistent detection performance is observed for the car class in ego scenarios. For frame-based detection, cars and motorcycles exhibit better performance than pedestrian detection. Notably, detectors show similar performance in noon and twilight conditions, with a decline observed in nighttime conditions for both fixed and ego perception settings. Figure~\ref{fig04} shows qualitative detection results for ego and fixed perception.

\begin{table}[t]
\centering
\resizebox{1\columnwidth}{!}{%
\begin{tabular}{c|ccc||ccc}
\hline \hline
\multirow{2}{*}{\textbf{Train Set}} & \multicolumn{3}{c||}{\textbf{Synthetic}}                & \multicolumn{3}{c}{\textbf{Real world}}               \\ \cline{2-7} 
 &
  \multicolumn{1}{c|}{\textbf{Car}} &
  \multicolumn{1}{c|}{\textbf{Pedestrian}} &
  \textbf{All Classes} &
  \multicolumn{1}{c|}{\textbf{Car}} &
  \multicolumn{1}{c|}{\textbf{Pedestrian}} &
  \textbf{All Classes} \\ \hline \hline
\textbf{Fixed}                        & \multicolumn{1}{c|}{0.537} & \multicolumn{1}{c|}{0.810} & 0.552  & \multicolumn{1}{c|}{0.384} & \multicolumn{1}{c|}{0.217} & 0.391 \\ \hline

\hline 
\end{tabular}%
}
\caption{Generalization assessment of RVT model, trained on synthetic data and evaluated on real-world fixed perception data.}
\label{table04}
\end{table}

\begin{figure}[t]
\centering
\includegraphics[width=1\linewidth]{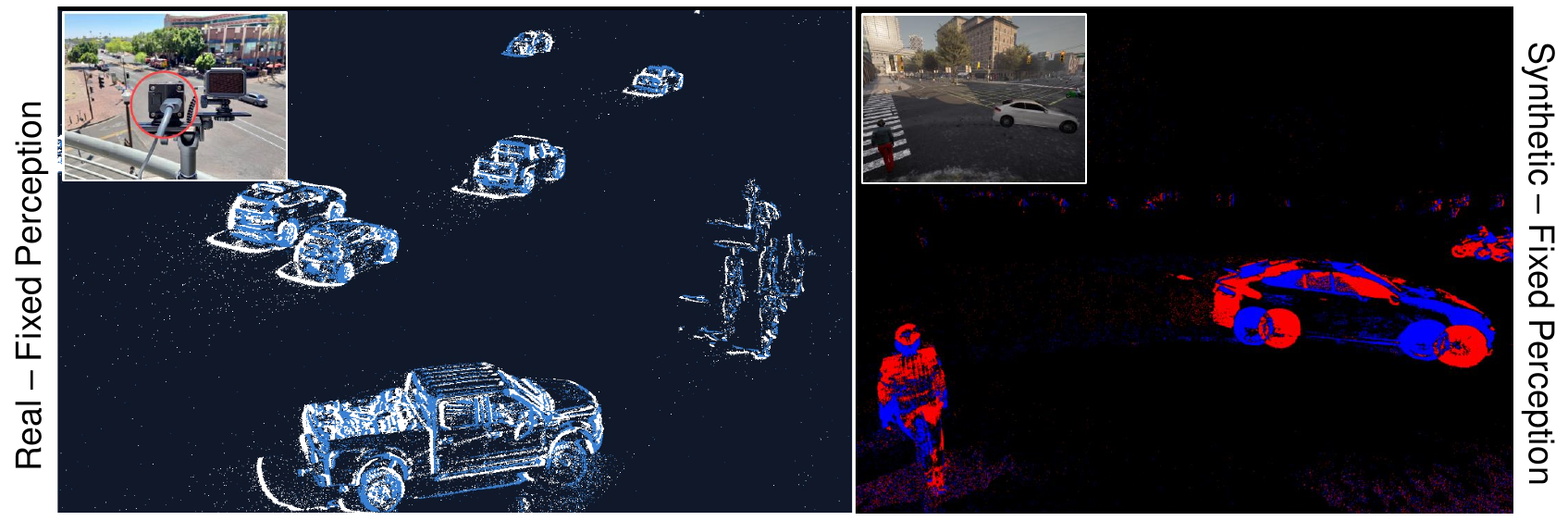}
\caption{Real-World Fixed Perception Event-Data Acquisition: Data captured at an intersection using the high-resolution Prophesee EVK4 HD event camera (left) similar to a setting used in CARLA (right).}
\label{fig05}
\end{figure}

\begin{figure*}[t]
\centering
\includegraphics[width=0.92\linewidth]{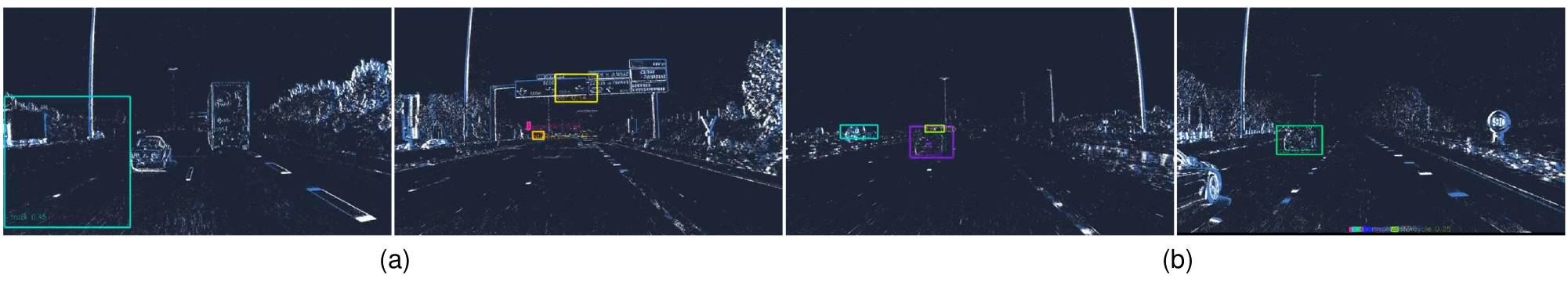}
\caption{Qualitative Results: Sample of detection instances in ego perception,  showing (a) incorrect and (b) correct detections, highlighting the variability in model performance.
}
\label{fig06}
\end{figure*}

\subsection{Generalization on  real-world data}

To assess the synthetic data-trained model's generalization capabilities on real-world data, we conduct quantitative and qualitative experiments on synthetic event-based detector models. For this purpose, we opt for RVT due to its superior performance compared to RED.

In our quantitative evaluation, we assess the fixed-perception detector model over real-world fixed event-based data, presenting the results in Table~\ref{table04}. The real-world event data was collected at an intersection near a university campus using the high-resolution Prophesee EVK4 HD event camera~\cite{prophesee-evk4}.
The event camera was strategically positioned at approximately $6\,\text{m}$ with a pitch angle of about $35^\circ$ to the ground, as shown in Figure~\ref{fig05} (left). This configuration mirrors that of DVS sensors in the CARLA simulation environment. RVT exhibits a relatively decent performance transitioning to real-world scenarios across all classes of objects, with a drop in performance for the pedestrian class.
We present qualitative results for ego perception, utilizing 1 Megapixel Automotive dataset from Prophesee \cite{perot2020learning} as depicted in Figure~\ref{fig06}. We intend to quantitatively evaluate real-world ego and fixed \cite{verma2024etram} perception scenarios as part of extended work. 

\section{Discussion}

In this section, we highlight the advantages of the \textit{SEVD} dataset across various domains. The high temporal resolution of event-based cameras significantly enhances performance in numerous tasks such as detection~\cite{perot2020learning, kugele2023many}, tracking ~\cite{zhang2022spiking, zhang2021object, lu2023virtual}, ReID ~\cite{ ahmad2023person, cao2023event}, trajectory prediction~\cite{monforte2023fast}, optical flow~\cite{nagata2023tangentially},  feature tracking ~\cite{messikommer2023data, seok2020robust}, and SLAM~\cite{jiao2021comparing, yang2019live}. This is particularly beneficial for autonomous driving applications, where real-time response is crucial as we offer dataset over $9\text{M}$ bounding box annotations ($2\text{D}$ and $3\text{D}$) in different formats (COCO, Pascal VOC, and KITTI), along with tracking IDs, facilitating tasks like $2\text{D}$/$3\text{D}$ detection, tracking, re-identification, and trajectory prediction.

\textit{SEVD}, being a multi-view synthetic vision-based ego and fixed perception dataset, plays a vital role in supporting existing methods that rely on data fusion from different sensor modalities to overcome the limitations of single sensor types and enhance detection performance. Further, multiview helps to overcome challenges, such as occlusion, limited perception horizon due to a restricted field of view, and low-point density at distant regions, posed by a single viewpoints.~\cite{arnold2020cooperative}.

Moreover, our dataset extends beyond ego-motion perception to include data from infrastructure perception, supporting research in Vehicle-to-Infrastructure (V2I) communication. This inclusion allows for the exploration of cooperative perception systems, where information sharing between vehicles and infrastructure enhances situational awareness and navigation safety~\cite{stevens2012benefits}. Such advancements can expedite the development of innovative solutions for V2I communication and collaborative autonomous systems, ultimately fostering safer and more efficient transportation networks.

\section{Conclusion}
The \textit{SEVD} dataset marks a notable advancement in synthetic event-based dataset for autonomous driving and traffic monitoring, offering $27\,\text{hr}$ of fixed and $31\,\text{hr}$ of ego perception event data, complemented by an equal amount of data from other sensor types. In total, the dataset encompasses a substantial $162\,\text{hr}$ of fixed and $186\,\text{hr}$ of ego perception data, offering a comprehensive view of diverse environmental conditions, including various lighting and weather scenarios across different scenes. This rich diversity of scenarios within \textit{SEVD} provides researchers with ample opportunities for exploration. The dataset's extensive annotation framework, featuring over $9\text{M}$ bounding boxes for various traffic participants in both $2\text{D}$ and $3\text{D}$ formats, along with raw event streams and their corresponding data from other sensor types, facilitates a deeper understanding of synthetic vision data and enables more effective algorithm development and evaluation. Furthermore, We report baselines for $2\text{D}$ detection tasks using state-of-the-art event-based detectors and detection performance for frame-based data. We believe \textit{SEVD} serves as a valuable resource for researchers and practitioners in the field, supporting advancements in event-based vision technology and contributing to the development of safer and more efficient transportation systems. 


{\small
\bibliographystyle{ieee_fullname}
\bibliography{PaperForReview}
}

\end{document}